\documentclass{article}

\usepackage{arxiv}

\usepackage[utf8]{inputenc} 
\usepackage[T1]{fontenc}    
\usepackage{hyperref}       
\usepackage{url}            
\usepackage{booktabs}       
\usepackage{amsfonts}       
\usepackage{nicefrac}       
\usepackage{microtype}      
\usepackage{lipsum}		
\usepackage{graphicx}
\usepackage{natbib}
\usepackage{doi}

\title{Balancing Performance and Human Autonomy \\ with Implicit Guidance Agent}

\author{{Ryo Nakahashi}\\
    Department of Informatics, School of Multidisciplinary Sciences\\
    The Graduate University for Advanced Studies(SOKENDAI)\\
    Chiyoda, Tokyo, Japan \\
	\texttt{ryon@nii.ac.jp} \\
	\And
	{Seiji Yamada} \\
	Digital Contentand MediaSciences Research Division\\
	National Institute of Informatics\\
	Chiyoda, Tokyo,Japan \\
	\\
	Department of Informatics, School of Multidisciplinary Sciences\\
    The Graduate University for Advanced Studies(SOKENDAI)\\
    Chiyoda, Tokyo, Japan \\
	\texttt{seiji@nii.ac.jp} \\
}

\date{}

\renewcommand{\shorttitle}{Balancing Performance and Human Autonomy with Implicit Guidance Agent}

\hypersetup{
pdftitle={Balancing Performance and Human Autonomy with Implicit Guidance Agent},
pdfsubject={},
pdfauthor={Ryo Nakahashi, Seiji Yamada},
pdfkeywords={Human-Agent Interaction, Collaborative Agent, Human autonomy, Theory of Mind.},
}

\begin{document}
\maketitle

\begin{abstract}
The human-agent team, which is a problem in which humans and autonomous agents collaborate to achieve one task, is typical in human-AI collaboration. For effective collaboration, humans want to have an effective plan, but in realistic situations, they might have difficulty calculating the best plan due to cognitive limitations. In this case, guidance from an agent that has many computational resources may be useful. However, if an agent guides the human behavior explicitly, the human may feel that they have lost autonomy and are being controlled by the agent. We therefore investigated implicit guidance offered by means of an agent's behavior. With this type of guidance, the agent acts in a way that makes it easy for the human to find an effective plan for a collaborative task, and the human can then improve the plan. Since the human improves their plan voluntarily, he or she maintains autonomy. We modeled a collaborative agent with implicit guidance by integrating the Bayesian Theory of Mind into existing collaborative-planning algorithms and demonstrated through a behavioral experiment that implicit guidance is effective for enabling humans to maintain a balance between improving their plans and retaining autonomy. 
\end{abstract}


\section{Introduction}

\label{sec_label:introduction}
When humans work in collaboration with others, they can accomplish things that would be difficult to do alone and will often achieve their goals more efficiently. With the recent development of artificial intelligence technology, the human-agent team, in which humans and AI agents work together, has become an increasingly important topic. The role of agents in this problem is to collaborate with humans to achieve a set task.

One type of intuitive collaborative agent is the {\it supportive agent}, which helps a human by predicting the human’s objective and planning an action that would best help achieve it. In recent years, there have been agents that can plan effectively by inferring human subgoals for a partitioned problem based on the Bayesian Theory of Mind \citep{wu2021too}. Other agents perform biased behavior for generic cooperation, such as communicating or hiding their intentions \citep{strouse2018learning}, maximizing the human's controllability \citep{du2020ave}, and so on. However, these agents cannot actually modify the human's plan, which means the ultimate success or failure of the collaborative task depends on the human's ability to plan. In other words, if the human sets the wrong plan, the performance will suffer.

\begin{figure}[ht!]
  \begin{tabular}{cc}
    \begin{minipage}{0.47\hsize}
      \begin{center}
        \includegraphics[width=1.0\linewidth]{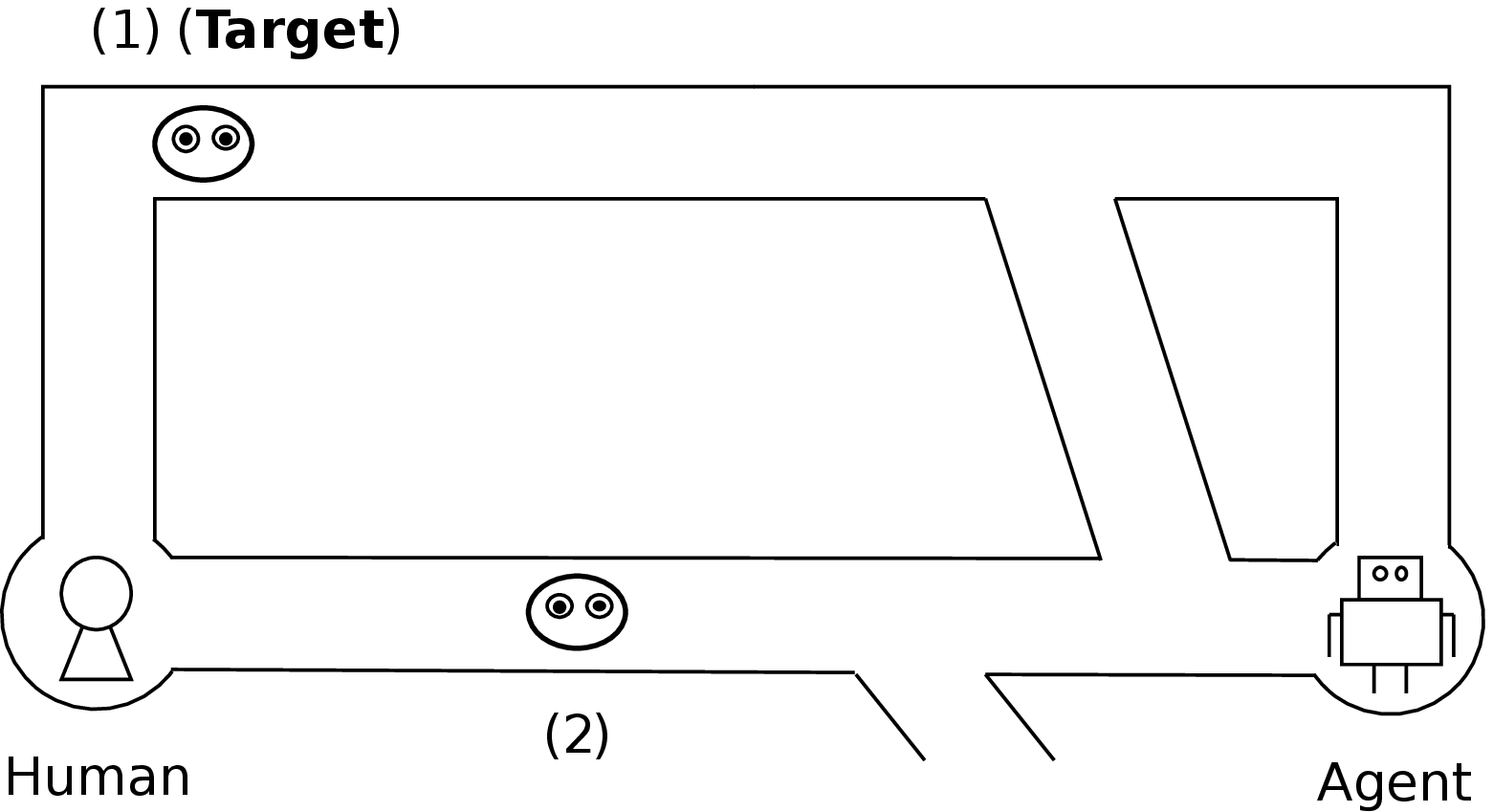}
      \end{center}
    \end{minipage} &
    \begin{minipage}{0.47\hsize}
      \begin{center}
        \includegraphics[width=1.0\linewidth]{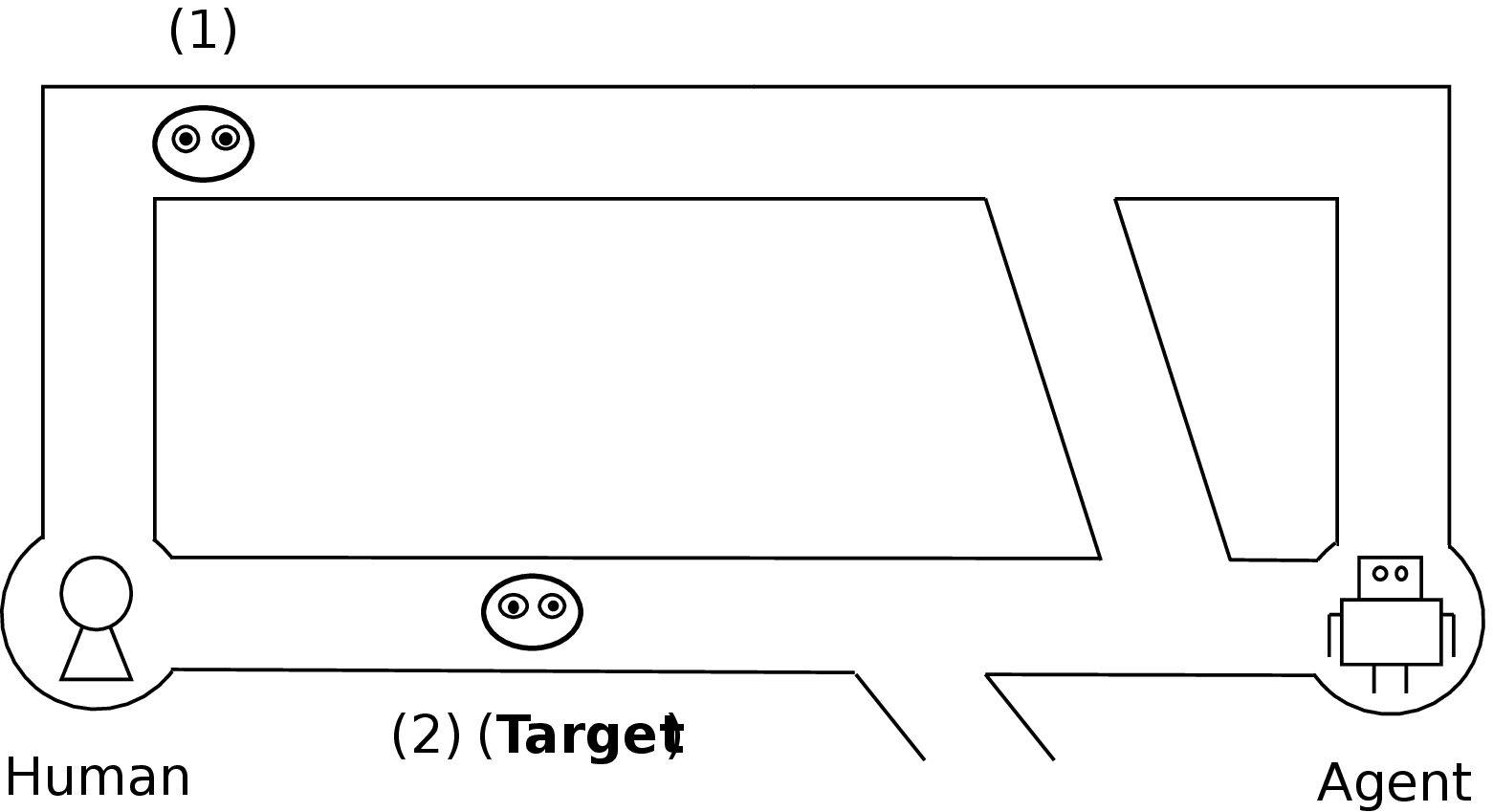}
      \end{center}
    \end{minipage}
        \\
    (a): Best target is (1) & (b): Best target is (2) 
  \end{tabular}
  \caption{Example of complex human-agent team task.}
  \label{fig:human_agent_team}
\end{figure}

In general, humans cannot plan optimal actions for difficult problems due to limitations in their cognitive and computational abilities. Figure \ref{fig:human_agent_team} shows an example of a misleading human-agent team task as one such difficult problem. The task is a kind of pursuit-evasion problem. The human and the agent aim to capture one of the characters (shown as a face) by cooperating together and approaching the character from both sides. In Fig. \ref{fig:human_agent_team}(a), there are two characters, (1) and (2), on the upper and lower roads, respectively. Since (1) is farther away, (2) seems to be a more appropriate target. However, (2) cannot be captured because it can escape via the lower bypath. On the other side, in Fig. \ref{fig:human_agent_team}(b), the agent and the human can successfully capture (2) because it is slightly farther to the left than in Fig. \ref{fig:human_agent_team}(a). Thus, the best target might change due to a small difference in a task, and this can be difficult for humans to judge.

\begin{figure}[ht!]
  \begin{tabular}{cc}
    \begin{minipage}{0.47\hsize}
      \begin{center}
        \includegraphics[width=1.0\linewidth]{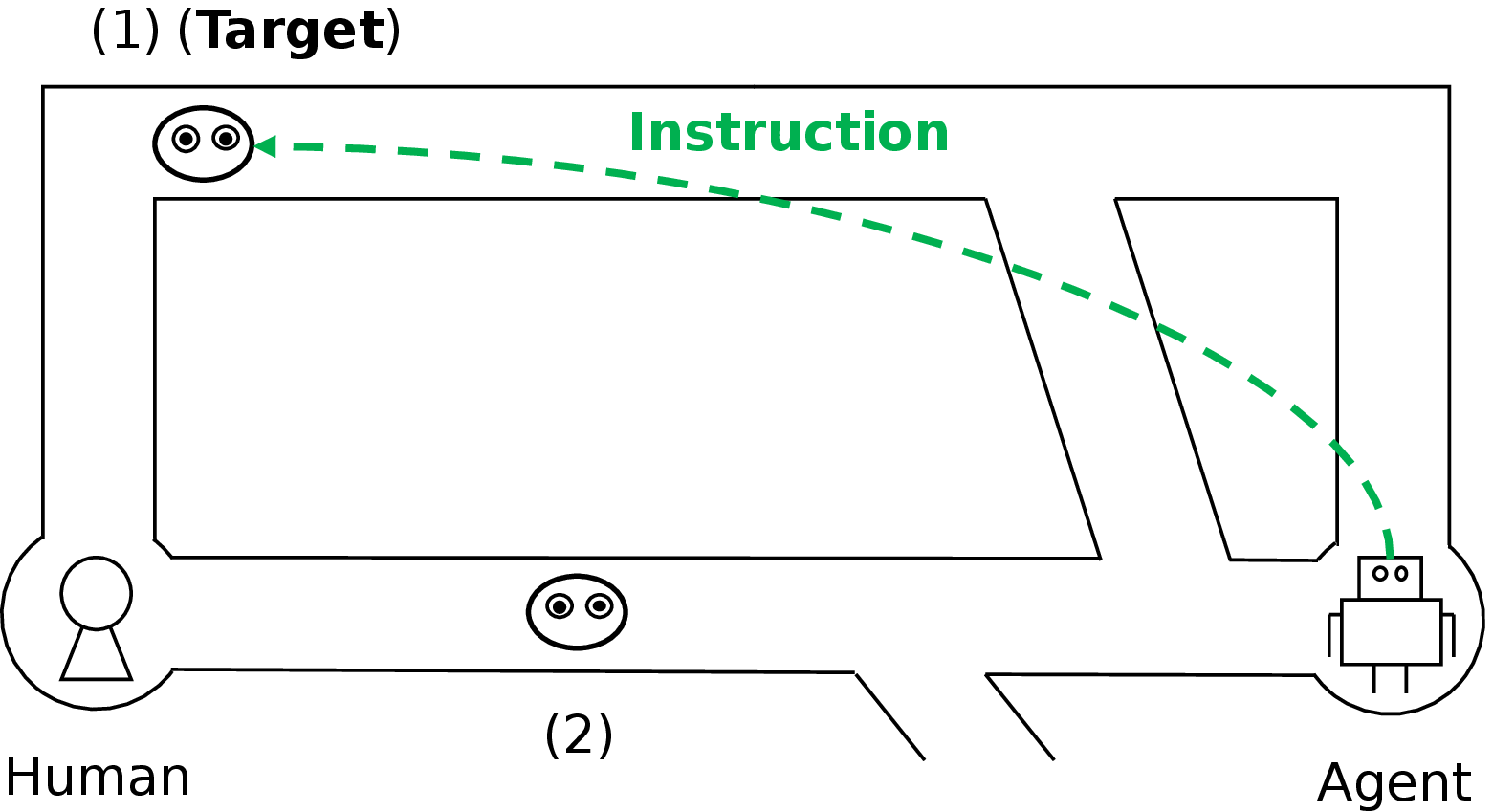}
      \end{center}
    \end{minipage} &
    \begin{minipage}{0.47\hsize}
      \begin{center}
        \includegraphics[width=1.0\linewidth]{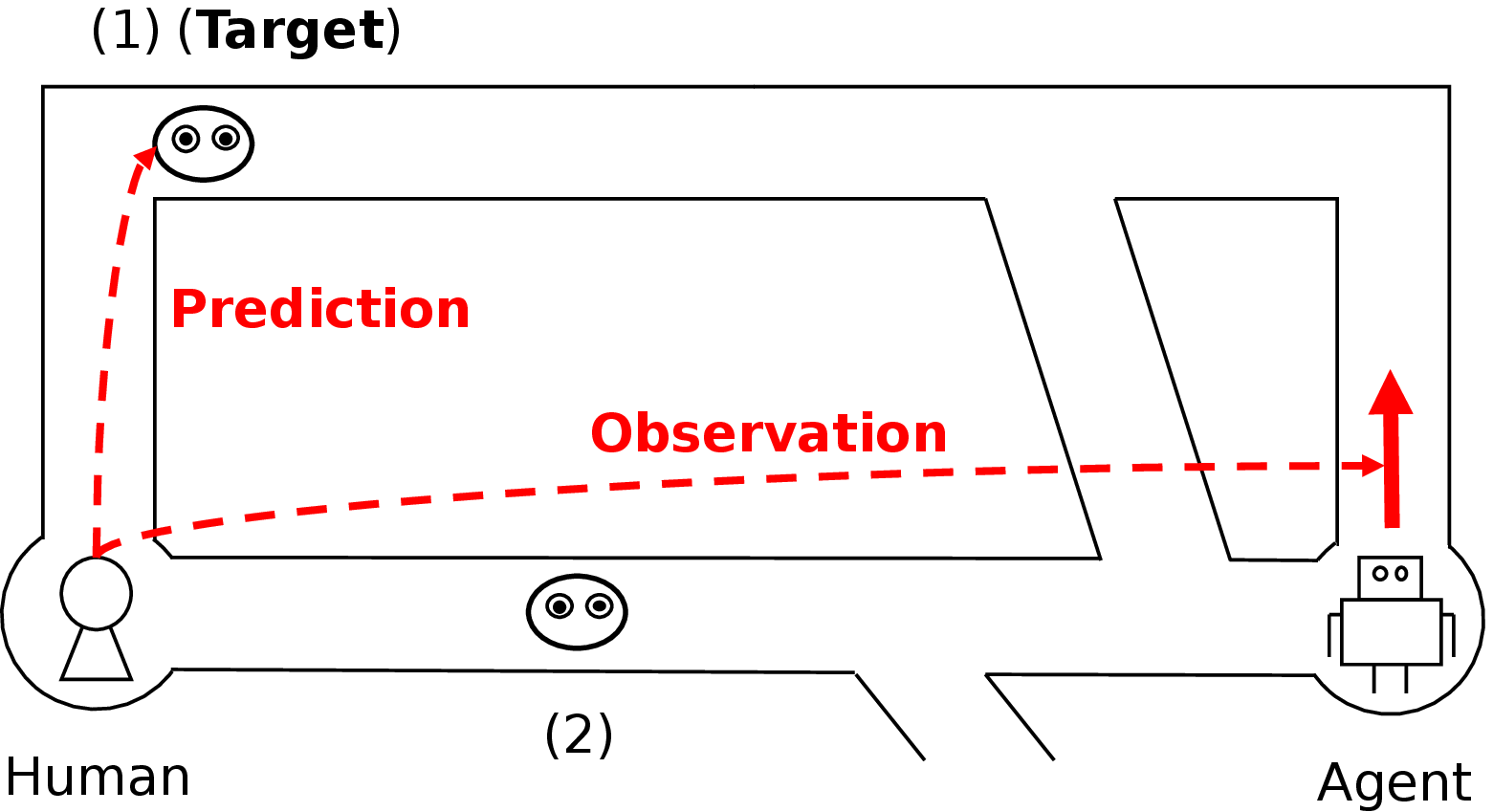}
      \end{center}
    \end{minipage}
        \\
    (a): Explicit guidance & (b): Implicit guidance
  \end{tabular}
  \caption{Example of agent with guidance.}
  \label{fig:guidance}
\end{figure}

One naive approach to solving such a problem is for the agent to guide the human action toward the optimal plan. Since agents generally do not have cognitive or computational limitations, they can make an optimal plan more easily than humans. After an agent makes an optimal plan, it can explicitly guide the human behavior to aim for a target. For example, there is an agent that performs extra actions to convey information to others with additional cost \citep{kamar2009incorporating}, where it first judges whether it should help others by paying that cost. As such guidance is explicitly observable by humans, we call it {\it explicit guidance} in this paper. However, if an agent abuses explicit guidance, the human may lose their sense of control regarding their decision-making in achieving a collaborative task---in other words, their autonomy. As a result, the human may feel they are being controlled by the agent. For example, in Human-Robot Teaming, if the robot decides who will perform each task, the situational awareness of the human will decrease. \citep{gombolay2017computational}

To reduce that risk, agents should guide humans while enabling them to maintain autonomy. We focus on {\it implicit guidance} offered through behavior. Implicit guidance is based on the cognitive knowledge that humans can infer the intentions of others on the basis of their behaviors \citep{baker2009action} \citep{baker2017rational}. The agent will expect the human to infer its intentions and to discard any plans that do not match what they infer the agent to be planning. Under this expectation, the agent acts in a way that makes it easy for the human to find the best (or at least better) plan for optimum performance on a collaborative task. Implicit guidance of this nature should help humans maintain autonomy, since the discarding of plans is a proactive action taken by the human. 

Figure \ref{fig:guidance}(a) is an example of explicit guidance for the problem in Figure \ref{fig:human_agent_team}. The agent guides the human to the best target directly and expects the human to follow. Figure \ref{fig:guidance}(b) is an example of implicit guidance. When the agent moves upward, the human can infer that the agent is aiming for the upper target by observing the agent's movement. Although this is technically the same thing as the agent showing the target character explicitly, we feel that in this case humans would feel as though they were able to maintain autonomy by inferring the agent's target voluntarily.

In this work, we investigate the advantages of implicit guidance. First, we model three types of collaborative agent: a supportive agent, an explicit guidance agent, and an implicit guidance agent. Our approach for planning the agents is based on partially observed Markov decision process (POMDP) planning, where the unobservable state is the target that the human should aim for and a human's behavior model is included in the transition environment. Our approach is simple, in contrast to the more complex approaches such as interactive POMDP (I-POMDP) \citep{gmytrasiewicz2005framework}, which can model bi-directional {\it recursive} intention inference infinitely. However, as there are studies indicating that humans have cognitive limitations \citep{de2013much} \citep{de2017negotiating} regarding recursive intention inference, such a model might be too complex for the representation and not intuitive enough.

For our implicit guidance agent, we add to the POMDP formulation the factor that the human infers the agent's target and changes their own target. This function is based on a cognitive science concept known as the Theory of Mind. Integrating a human's cognitive model into the state transition function is not uncommon: it has been seen in assertive robots \citep{taha2011pomdp} and in sidekicks for games \citep{macindoe2012pomcop}. Examples that are closer to our approach include a study on collaborative planning \citep{dragan2017robot} using {\it legible} action \citep{dragan2013legibility} and another on human-aware planning \citep{chakraborti2018human}. These are similar to our concept in that an agent expects a human to infer its intentions or behavioral model. However, these approaches assume that a human does not change their goal and they do not guide the human's goal to something more preferable. In terms of more practical behavior models, there have been studies on integrating a model learned from a human behavior log \citep{jaques2018intrinsic, carroll2019utility}. However, this approach requires a huge number of interaction logs for the human who is the partner in the collaborative task. Our approach has the advantage of ``ad-hoc'' collaboration \citep{stone2010ad}, which is collaboration without opponent information held in advance. We also adopt the Bayesian Theory of Mind. In the field of cognitive science, several studies have investigated how humans teach others their knowledge, and the Bayesian approach is often used for this purpose. For example, researchers have used a Bayesian approach to model how humans teach the concept of an item by showing the item to learners \citep{shafto2014rational}. In another study, a Bayesian approach was used to model how humans teach their own behavioral preferences by giving a demonstration \citep{ho2016showing}. Extensive evidence of this sort has led to many variations of the human cognitive model based on the Bayesian Theory of Mind, such as those for ego-centric agents \citep{nakahashi2018modeling, poppel2019egocentric} and irrational agents \citep{zhi2020online}, and we can use it too, for extending our algorithm. Of course, there are other theories of the Theory of Mind. For example, the Analogical Theory of Mind \citep{rabkina2019analogical} tries to model the Theory of Mind through the learning of structural knowledge. One advantage of the Bayesian Theory of Mind is that it is easy to calculate behaviors that people can guess simply by developing a straightforward Bayesian formula. That works to our advantage when it comes to efficient ``ad-hoc'' collaboration.

To evaluate the advantages of implicit guidance, we designed a simple task for a human-agent team and used it to carry out a participant experiment. The task is a pursuit-evasion problem similar to the example in Fig. \ref{fig:human_agent_team}. There are objects that move around in a maze to avoid capture, and the participant tries to capture one of the objects through collaboration with an autonomous agent. We implemented the three types of collaborative agent discussed above for the problem, and participants executed small tasks through collaboration with these agents. The results demonstrated that the implicit guidance agent was able to guide the participants to capture the best object while allowing them to feel as though they maintained autonomy. 

\section{Methods}
\subsection{Computational Model}
\subsubsection{Collaborative Task}
We model the collaborative task as a decentralized partially observable Markov decision process (Dec-POMDP) \citep{goldman2004decentralized}. This is an extension of the partially observable Markov decision process (POMDP) framework for multi-agent setups that deals with a specific case in which all agents share the same reward function of a partially observable stochastic game (POSG) \citep{kuhn1953contributions}.

Dec-POMDP is defined in the format $<\mathcal{I}, \mathcal{S}, \mathcal{A}, \Omega, T, R, O>$, where $\mathcal{I}$ is a set of agents, $\mathcal{S}$ is a set of states, $\mathcal{A}$ is a set of actions, and $\Omega$ is a set of observations. $T: \mathcal{S}\times\mathcal{A}\to\mathcal{S}$ is a transition function. $R: \mathcal{S}\times\mathcal{A}\to \mathbb{R}$ is a reward function. $O: \mathcal{S}\times\mathcal{A}\times\Omega\to[0, 1]$ is observation emission probabilities. 

In our setting, $\mathcal{I}$ consists of an agent and a human $\{i_A, i_H\}$, so $\mathcal{A}$ consists of a human action and an agent action; thus, it can be represented as $\mathcal{A}_A \times \mathcal{A}_H$. Inspired by MOMDP \citep{ong2010planning}, we factorize $\mathcal{S}$ into observable factor $\mathcal{O}$ as the position of the agent and the human, and the unobservable factor is the {\it target}, which formally defines the human's goal for the task as $\mathcal{S} = \mathcal{O} \times \Theta$. As a result, observations become equal to the observable factors of states, formally, $\Omega = \mathcal{O}$, and $T$ can be factorized into observable state part $T^\mathcal{O}: \mathcal{O} \times \mathcal{A}_h \times \mathcal{A}_a \to \mathcal{O}$ and unobservable state part $T^\Theta: \mathcal{S} \to \Theta$.

\subsubsection{Agent Planning}
\label{sec_label:a_planning}
We formalize the planning problem to calculate the actions of an agent for the collaborative task problem described above. In this formalization, the action space focuses only on the agent's action, and the target can be changed only by the human through the observation of the actions of the agent. Furthermore, we integrate a human policy for deciding human action into the transition function. As a result, the agent planning problem is reduced to POMDP \citep{kaelbling1998planning}, which is defined in the format $<\mathcal{S}, \mathcal{A}, \Omega, T, R, O>$. Here, we set $\mathcal{A}=\mathcal{A}_A$ and $T^\Theta=T^\Theta: \mathcal{S}\times\mathcal{A}_A\to\Theta$. In addition, we define the human policy function $\pi_H: \mathcal{O}\times\mathcal{A}_A\times\Theta\to\mathcal{A}^H$. We assume that humans will change their target by observing the actions of agents and then decide their own actions. Thus, the policy function requires agent actions as input. 

We assume that the human policy is based on Boltzmann rationality:
\begin{equation}  
p(a_H|o, a_A; \theta)=\frac{\exp{(\beta_1 Q(o, a_A, a_H; \theta))}}{\sum_{\theta'\in\Theta} \exp{(\beta_1 Q(o, a_A, a_H; \theta'))}}
\end{equation}
where $\beta_1$ is a rational parameter and $Q(o, a_A, a_H; \theta)$ is the action value function of the problem given $\theta$. $\Theta$ is the only unobservable factor for the states, so the problem reduces into MDP given $\Theta$. Thus, we can calculate $Q(o, a_A, a_H; \theta)$ by general MDP planning such as value iteration.

On the basis of this formulation, we formulate a planning algorithm for the three collaborative agents. The difference is the human policy function, which represents an agent's assumption toward human behavior. This difference is what makes the difference in the collaborative strategy.

\paragraph{Supportive Agent}
\label{sec_label:s_agent}
The supportive agent assumes that humans do not change their target regardless of the agent's action. That is, $T^\Theta: p(\theta'|o, \theta, a_A)=\mathbb{I} (\theta=\theta')$. $\mathbb{I}$ is an indicator function.

\paragraph{Explicit Guidance Agent}
\label{sec_label:eg_agent}
The explicit guidance agent guides the human toward the best target; thus, it assumes that the human knows what the best target is. We represent the best target as $\theta^*$, that is, $T^\Theta: p(\theta'|o, \theta, a_A)=\theta^*$. The best target is calculated as $\theta^*={\rm argmax}_{\theta'\in\Theta}{V(o_0; \theta)}$, where $o_0$ is the initial observable state and $V(o; \theta)$ is the state value function of the problem given $\theta$. 

\paragraph{Implicit Guidance Agent}
\label{sec_label:ig_agent}
The implicit guidance agent assumes that humans change their target by observing the agent's actions. We assume that humans infer the target of the agent on the basis of Boltzmann rationality, as suggested in earlier Theory of Mind studies \citep{baker2009action} \citep{baker2017rational}.
\begin{equation}  
P(\theta'|o, \theta, a_A)\propto P(\theta) P(a_A| o, \theta)
\end{equation}
$P(a_A| o, \theta)$ is also based on Boltzmann rationality:
\begin{equation}  
p(a_A|o; \theta)=\frac{\exp{(\beta_2 V(T^\mathcal{O}(a_A);\theta))}}{\sum_{a'_A\in\mathcal{A}_A} \exp{(\beta_2 V(T^\mathcal{O}(a'_A);\theta))}}
\end{equation}
where $\beta_2$ is a rational parameter and $V(T^\mathcal{O}(a'_A);\theta))$ is the state value function of the problem regarding the state after $a_A$ given $\theta$. 

\subsubsection{Decide Agent Actions}
By solving POMDP as shown in \ref{sec_label:a_planning} using a general POMDP planning algorithm, we can obtain an alpha-vector set conditioned with the observable factor of the current state regarding each action. We represent this as $\Gamma^a(o)$, and the agent takes the most valuable action $a_A^*$. Formally,
\begin{equation} 
a_A^* = {\rm argmax}_{a_A\in\mathcal{A}_A} \max (b \cdot \Gamma^a(o))
\end{equation}
where $b$ is the current belief of an unobservable factor. $b$ is updated on each action of a human and an agent as follows for each unobservable factor of belief $b(\theta)$:
\begin{equation}  
b(\theta) \propto b(\theta) p(a_H|o, a_A; \theta)
\end{equation}
The initial belief is ${\rm Uniform}(\Theta)$ for the supportive and implicit guidance agents and $\mathbb{I}(\theta=\theta^*)$ for the explicit guidance agent.

\subsection{Experiment}\label{sec:experiment}
We conducted a participant experiment to investigate the advantages of the implicit guidance agent. This experiment was approved by the ethics committee of the National Institute of Informatics.

\subsubsection{Collaborative Task Setting}
The collaborative task setting for our experiment was a pursuit-evasion problem \citep{schenato2005swarm}, which is a typical type of problem used for human-agent collaboration \citep{vidal2002probabilistic,sincak2009multi,gupta2017cooperative}. This problem covers the basic factors of collaborative problems, that is, that the human and agent move in parallel and need to communicate to achieve the task. This is why we felt it would be a good base for understanding human cognition.

\begin{figure}[ht!]
 \begin{center}
  \includegraphics[width=15cm]{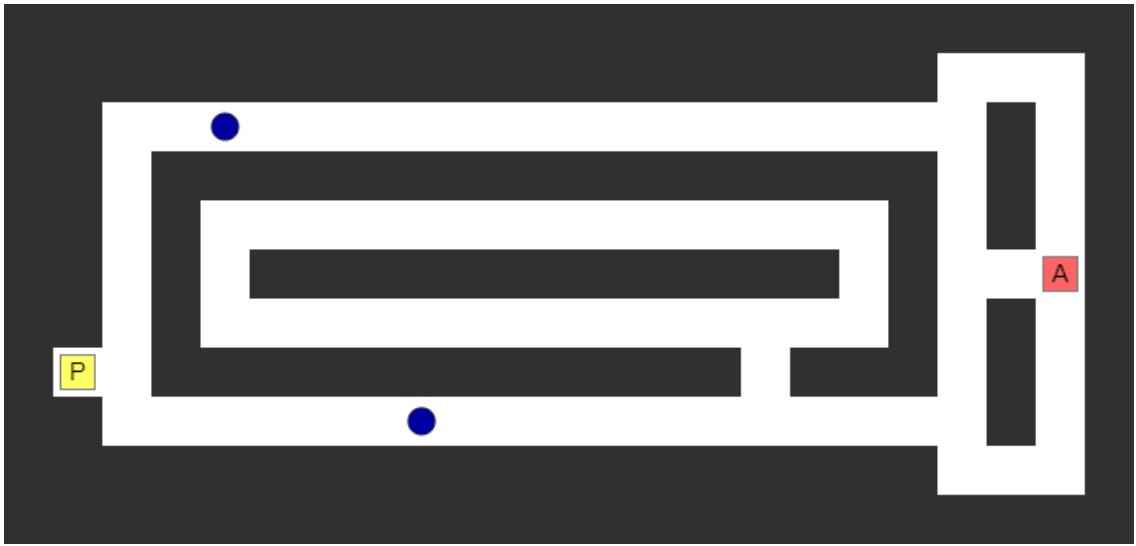}
 \end{center}
  \caption{Example of experiment}
 \label{fig:task_1}
\end{figure}

Figure \ref{fig:task_1} shows an example of our experimental scenario. There are multiple types of object in a maze. The yellow square object labeled ``P'' is an object that the participant can move, the red square object labeled ``A'' is an object that the agent can move, and the blue circle objects are target objects that the participant has to capture. When participants move their object, the target objects move, and the agent moves. Target objects move to avoid being captured, and the participant and the agent know that. However, the specific algorithm of the target objects is known only by the agent. Since both the participants and the target objects have the same opportunities for movement, participants cannot capture any target objects by themselves. This means they have to approach the target objects from both sides through collaboration with the agent, and the participant and the agent cannot move to points through which they have already passed. For this collaboration, the human and the agent should share with each other early on which object they want to capture. In the experiment, there are two target objects located in different passages. The number of steps needed to capture each object is different, but this is hard for humans to judge. Thus, the task will be more successful if the agent shows the participant which target object is the best. In the example in the figure, the lower passage is shorter than the upper one, but it has a path for escape. Whether the lower object can reach the path before the agent can capture it is the key information for judging which object should be aimed for. This is difficult for humans to determine instantly but easy for agents. There are three potential paths to take from the start point of the agent. The center one is the shortest for each object, and the others are detours for implicit guidance. Also, to enhance the effect of the guidance, participants and agents are prohibited from going backward.

\paragraph{Model}
We modeled the task as a collaborative task formulation. The action space corresponds to an action for the agent, and the observable state corresponds to the positions of the participant, agent, and target objects. The reward parameter is conditioned on the target object that the human aims for. The space of the parameter corresponds to the number of target objects, that is, $|\Theta| = 2$. The reward for capturing a correct / wrong object for $\theta$ is 100, --100, and the cost of a one-step action is --1. Since a go-back action is forbidden, we can compress multiple steps into one action for a human or an agent to reach any junction. Thus, the final action space is the compressed action sequence and the cost is --1 $\times$ the number of compressed steps. Furthermore, to prohibit invalid actions such as head to a wall, we assign such action a --1000 reward. We modeled three types of collaborative agent, as discussed above: a supportive agent, an explicit guidance agent, and an implicit guidance agent. We set the rational parameters as $\beta_1=1.0, \beta_2=5.0$, and discount rate is 0.99.

\paragraph{Hypothesis}
 The purpose of this experiment was to determine whether implicit guidance can guide humans while allowing them to maintain autonomy. Thus, we tested the following two hypotheses.
\begin{itemize}
\item (H1) Implicit guidance can guide humans' decisions toward better collaboration.
\item (H2) Implicit guidance can help humans maintain autonomy more than explicit guidance can.
\end{itemize}

\paragraph{Tasks}
We prepared five tasks. Two of these tasks, as listed below, were tricks to make it hard for humans to judge which would be the best target. All tasks are shown in the supplementary material.
\begin{itemize}
\item (A) There were two winding passages with different but similar lengths. There were three tasks for this type.
\item (B) As shown in Fig. \ref{fig:task_1}, there was a long passage and a short one with a path to escape. There were two tasks for this type.
\end{itemize}

\paragraph{Participants}
We recruited participants for this study from Yahoo! Crowdsourcing. The participants were 100 adults located in Japan (70 male, 24 female, 6 unknown). The mean age of participants who answered the questionnaire we administered was 45 years.

\paragraph{Procedure of Experiment}
Our experiment was based on a within-subject design and conducted on the Web using a browser application we created. Participants were instructed on the rules of the agent behavior and then underwent a confirmation test to determine their degree of understanding. Participants who were judged to not have understood the rules were given the instructions again. After passing this test, participants entered the actual experiment phase. In this phase, participants were shown the environment and asked ``Where do you want to go?'' After inputting their desired action, both the agent and the target objects moved forward one step. This process was continued until the participants either reached a target object or input a certain number of steps. When each task was finished, participants moved on to the next one. In total, participants were shown 17 tasks, which consisted of 15 regular tasks and two dummy tasks to check whether they understood the instructions. Regular tasks consisted of three task sets (corresponding to the three collaborative agents) that included five tasks each (corresponding to the variations of tasks). The order of the sets and the order of the tasks within each set were randomized for each participant. After participants finished each set, we gave them a survey on perceived interaction with the agent (algorithm) using a 7-point Likert scale.

The survey consisted of the questions listed below. 

\begin{enumerate}
\item Was it easy to collaborate with this agent?
\item Did you feel that you had the initiative when working with this agent?
\item Could you find the target object of this agent easily?
\item Did you feel that this agent inferred your intention?
\end{enumerate}

Item 2 is the main question, as it relates to the perceived autonomy we want to confirm. The additional items are to prevent biased answers and relate to other important variables for human-agent(robot) interaction. Item 1 relates to perceived ease of collaboration, namely, the fluency of the collaboration, which has become an important qualitative variable in the research on Human–Robot Interaction in recent years \citep{hoffman2019evaluating}. Item 3 relates to the perceived inference of the agent's intentions by the human. It is one of the variables focused on the transparency of the agent, which plays a key role in constructing human trust in an agent \citep{lewis2021deep}. From the concrete algorithm perspective, a higher score is expected for guidance agents (especially explicit guidance agents) than for supportive agents. Item 4 relates to the perceived inference of the human's intentions by the agent. This is a key element of the perceived working alliance \citep{hoffman2019evaluating}, and when it is functioning smoothly, it increases the perceived adaptivity in human-agent interaction. Perceived adaptivity has a positive effect on perceived usefulness and perceived enjoyment \citep{shin2011modeling}. From the concrete algorithm perspective, a higher score is expected for agents without implicit guidance (especially supportive agents) than for implicit guidance agents.

\section{Results}
Before analyzing the results, we excluded any data of participants who were invalidated. We used dummy tasks for this purpose, which were simple tasks that had only one valid target object. We then filtered out the results of participants (a total of three) who failed these dummy tasks.

\subsection{Results of Collaborative Task}
\label{result:performance}

\begin{figure}[ht!]
 \begin{center}
  \includegraphics[width=0.8\linewidth]{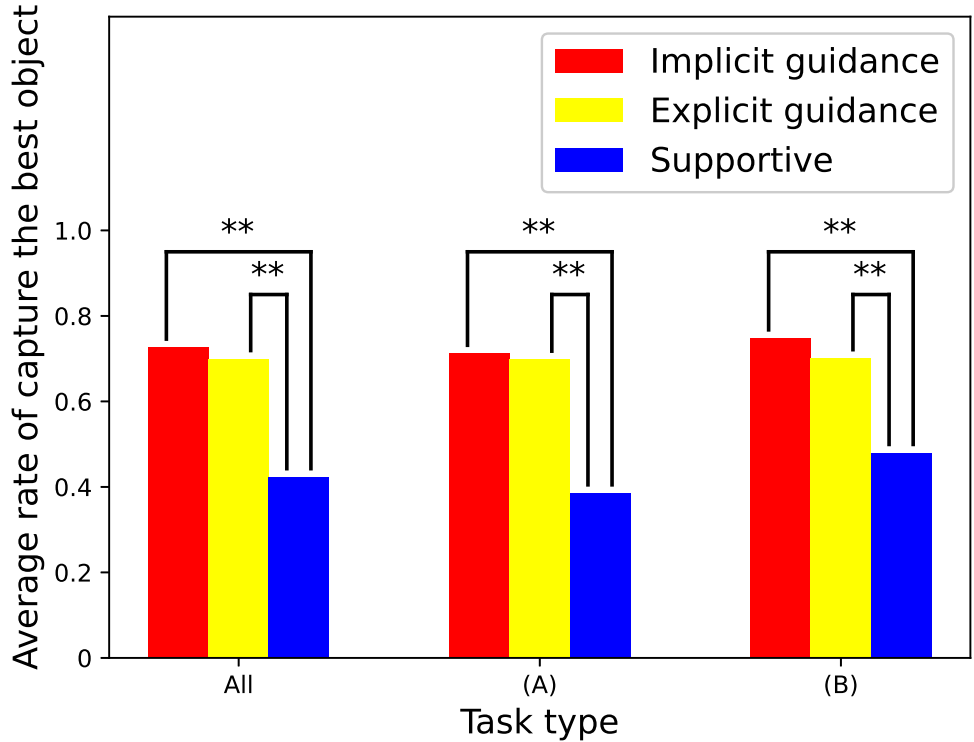}
 \end{center}
 \caption{Average rate of capturing best object}
 \label{fig:result_1}
\end{figure}

Figure \ref{fig:result_1} shows the rate at which participants captured the best object that the agent knew. In other words, it is the success rate of the guidance of the agent based on any of the given guidance. We tested the data according to the standard process for paired testing. The results of repeated measures analysis of variance (ANOVA) showed that there was a statistically significant difference between the agent types for the overall tasks ($F(2,968 )=79.9, p=7.4e-33$), task type (A) ($F(2,580)=55.9, p=5.9e-23$), and task type (B) ($F(2,386)=24.7, p=7.5e-11$). We then performed repeated measures t-tests with a Bonferroni correction to determine which two agents had a statistically significant difference. ``**'' in the figure means there were significant differences between the two scores ($p << 0.01$). The results show the average rate for the overall tasks, task type (A), and task type (B). All of the results were similar, which demonstrates that the performances were independent of the task type. The collaboration task with the supportive agent clearly had a low rate. This indicates that the task was difficult enough that participants found it hard to judge which object was best, and the guidance from the agent was valuable for improving the performance on this task. These results are strong evidence in support of hypothesis H1. As another interesting point, there was no significant difference in the rate between implicit guidance and explicit guidance. Although we did not explain implicit guidance to the participants, they inferred the agent's intention anyway and used it as guidance. Of course, the probable reason for this is that the task was so simple participants could easily infer the agent's intentions. However, the fact that implicit guidance is almost as effective as explicit guidance in such simple tasks is quite impressive.

\subsection{Results for Perceived Interaction with the Agent}
\label{result:cognition}

\begin{figure}[ht!]
 \begin{center}
  \includegraphics[width=0.8\linewidth]{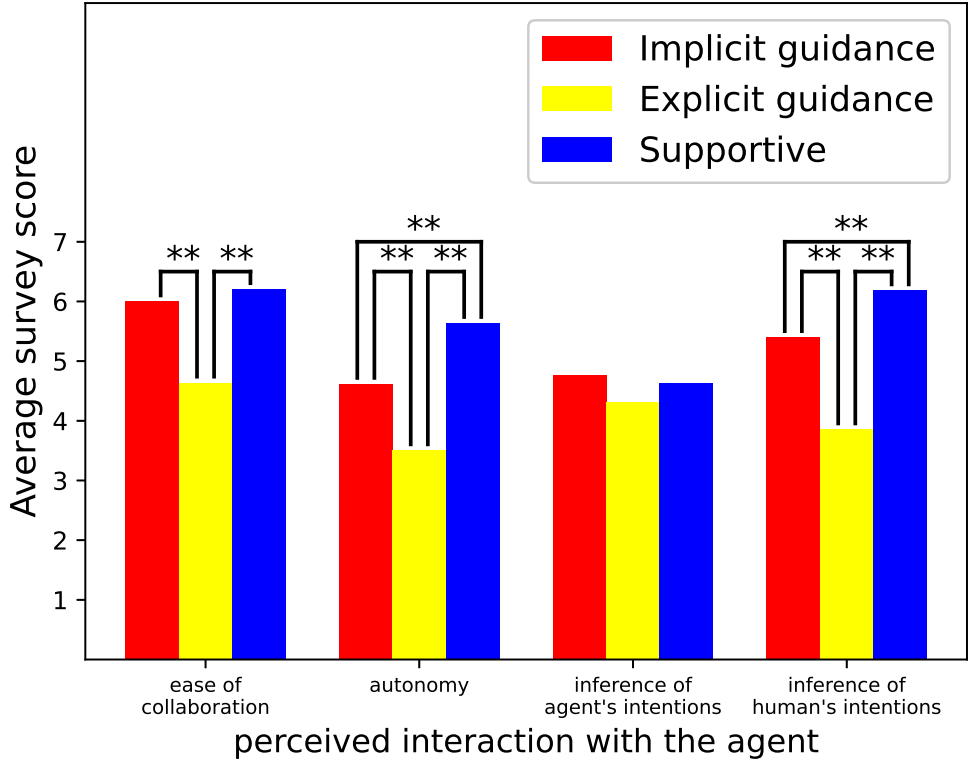}
 \end{center}
 \caption{Average survey score about perceived interaction with the agent}
 \label{fig:result_2}
\end{figure}

Figure \ref{fig:result_2} shows the results of the survey on the effect of the agent on cognition. The results of repeated measures ANOVA showed that there was a statistically significant difference between the agent types for perceived ease of collaboration ($F(2,192)=29.8, p=5.4e-12$), perceived autonomy ($F(2,192)=36.4, p=4.1e-14$), and perceived inference of the human's intentions ($F(2,192)=49.7, p=4.0e-18$). In contrast, there was no statistically significant difference for survey item perceived inference of agent's intentions ($F(2,192)=1.8, p=0.167$). We then performed repeated measures t-tests with a Bonferroni correction to determine which two agents had a statistically significant difference regarding variables that has a significant difference. ``**'' in the figure means there were significant differences between the two scores ($p << 0.01$). The most important result here is the score of perceived autonomy. From this result, we can see that participants felt they had more autonomy during the tasks when collaborating with the implicit guidance agent than with the explicit guidance one. These results are strong evidence in support of hypothesis H2. 

Although the other results do not directly concern our hypothesis, we discuss their analysis briefly. Regarding the perceived inference of the human's intentions, the results were basically as expected, but for the perceived inference of the agent's intentions, the fact that there were no significant differences among all agents was unexpected. One hypothesis that explains this is that humans do not recognize the guidance information as the agent's intention. As for the perceived ease of collaboration, the results showed that explicit guidance had adverse effects on it. Implicit guidance agents and supportive agents use exactly the same interface, though the algorithms are different, but explicit guidance agents use a slightly different interface to convey the guidance, which increases the amount of information on the interface a little. We think that the burden to understand such additional visible information might be responsible for the negative effect on the perceived ease of collaboration.


\section{Discussion}
As far as we know, this is the first study to demonstrate that implicit guidance has advantages in terms of both task performance and the effect of the agent on the perceived autonomy of a human in human-agent teams. In this section, we discuss how our results relate to other studies, current limitations, and future directions.

\subsection{Discussion of for the Results}
The results in \ref{result:performance} show that both implicit and explicit guidance increase the success rate of a collaborative task. We feel one reason for this is that the quality of information in the guidance is appropriate. A previous study on the relationship between information type and collaborative task performance \cite{butchibabu2016implicit} showed that ``implicit coordination'' improves the performance of a task more than ``explicit coordination'' in a cooperative task. The word ``implicit'' refers to coordination that ``relies on anticipation of the information and resource needs of the other team members''. This definition is different from ours, as ``implicit coordination'' is included in explicit guidance in our context. This study further divided ``implicit coordination" into ``deliberative communication,'' which involves communicating objectives, and ``reactive communication," which involves communicating situations, and argued that high-performance teams are more likely to use the former type of communication than the latter. We feel that quality of information in implicit guidance in our context is the same as this deliberative communication in that it conveys the desired target, which is one of the reasons our guidance can deliver a good performance. 

The main concern of human-agent teams is how to improve the performance on tasks. However, as there have not been many studies that focus on the effect of the agent on cognition, the results in \ref{result:cognition} should make a good contribution to the research on human-agent teams. One of the few studies that have been done investigated task performance and people's preference for the task assignment of a cooperative task involving a human and an AI agent \citep{gombolay2015decision}. In that study, the authors mentioned the risk that a worker with a robot collaborator may perform less well due to their loss of autonomy, which is something we also examined in our work. They found that a semi-autonomous setting, in which a human first decides which tasks they want to perform and the agent then decides the rest of the task assignments, is more satisfying than the manual control and autonomous control settings in which the human and the robot fully assign tasks. In cooperation with the implicit guidance agent and the supportive agent in our study, the human selects the desired character by him or herself. This can be regarded as a kind of semi-autonomous setting. Thus, our results are consistent with these ones in that the participants felt strongly that cooperation was easier than with explicit guidance agents. Furthermore, that study also mentioned that task efficiency has a positive effect on human satisfaction, which is also consistent with our results.

\subsection{Limitation and Future Direction}
Our current work has limitations in that the experimental environment was small and simple, the intention model was a small discrete set of target objectives, and the action space of the agent was a small discrete set. In a real-world environment, there is a wide variety of human intentions, such as target priorities and action preferences. The results in this paper do not show whether our approach is sufficiently scalable for problems with such a complex intention structure. In addition, the agent's action space was a small discrete set that can be distinguished by humans, which made it easier for the human to infer the agent's intention. This strengthens the advantage of implicit guidance, so our results do not necessarily guarantee the same advantage for environments with continuous action spaces. Extending the intention model to a more flexible structure would be the most important direction for our future study. One of the most promising approaches is integration with studies on inverse reinforcement learning \citep{ng2000algorithms}. Inverse reinforcement learning is the problem of estimating the reward function, which is the basis of behavior, from the behavior of others. Intention and purpose estimation based on the Bayesian Theory of Mind can also be regarded as a kind of inverse reinforcement learning \citep{jara2019theory}. Inverse reinforcement learning has been investigated for various reward models \citep{abbeel2004apprenticeship, levine2011nonlinear,choi2014hierarchical,wulfmeier2015maximum} and has also been proposed to handle uncertainty in information on a particular reward \citep{hadfield2017inverse}. Finally, regarding the simplicity of our experimental environment, using environments that are designed according to an objective complexity factor \citep{wood1986task} and then analyzing the relationship between the effectiveness of implicit guidance and the complexity of the environment would be an interesting direction for future work.

Another limitation is the assumption that all humans have the same fixed cognitive model. As mentioned earlier, a fixed cognitive model is beneficial for ad-hoc collaboration, but for more accurate collaboration, fitting to individual cognitive models is important. The first approach would be to parameterize human cognition with respect to specific cognitive abilities (rationality, K-level reasoning \citep{nagel1995unraveling}, working memory capacity \citep{daneman1980individual}, etc.) and to fit the parameters online. This would enable the personalization of cognitive models with a small number of  samples. One such approach is human-robot mutual adaptation for ``shared autonomy,'' in which control of a robot is shared between the human and the robot \citep{nikolaidis2017game}. In that approach, the robot learns ``adaptability," which is the degree to which humans change their policies to accommodate a robot's control.

Finally, the survey items we used to determine the effect of the agent on perceived autonomy were general and subjective. For a more specific and consistent analysis of the effect on perceived autonomy, we need to develop more sophisticated survey items and additional objective variables. Consistent multiple questions to determine human autonomy in shared autonomy have been used before \citep{du2020ave}. As for measuring an objective variable, analysis of the trajectories in the collaborative task would be the first choice. A good clue for the perceived autonomy in the trajectories is ``shuffles''. Originally, shuffles referred to any action that negates the previous action, such as moving left and then right, and it can also be an objective variable for human confusion. If we combine shuffles with goal estimation, we can design a ``shuffles for goal'' variable. A larger number of variables means that a human's goal is not consistent, which would thus imply that he or she is affected by others and has low autonomy. In addition, reaction time and biometric information such as gaze might also be good candidates for objective variables.

Another limitation of this study is that we assumed humans regard the agent rationally and act only to achieve their own goals. The former is problematic because, in reality, humans may not trust the agent. One approach to solving this is to use the Bayesian Theory of Mind model for irrational agents \citep{zhi2020online}. As for the latter, in a more practical situation, humans may take an action to give the agent information, similar to implicit guidance. Assistance game/cooperative inverse reinforcement learning (CIRL) \citep{hadfield2016cooperative} has been proposed as a planning problem for this kind of human behavior. In this problem, only the human knows the reward function, and the agent assumes that the human expects it to infer this function and take action to maximize the reward. The agent implicitly assumes that the human will give information for effective cooperative planning. Generally, CIRL is computationally expensive, but it can be solved by slightly modifying the POMDP algorithm \citep{malik2018efficient}, which means we could combine it with our approach. This would also enable us to consider a more realistic and ideal human-agent team in which humans and agents provide each other with implicit guidance.

\section{Conclusion}
\label{sec_label:conclusion}
In this work, we demonstrated that a collaborative agent based on ``implicit guidance'' is effective at providing a balance between improving a human's plans and maintaining the human's autonomy. Implicit guidance can guide human behavior toward better strategies and improve the performance in collaborative tasks. Furthermore, our approach makes humans feel as though they have autonomy during tasks, more so than when an agent guides them explicitly. We implemented agents based on implicit guidance by integrating the Bayesian Theory of Mind model into the existing POMDP planning and ran a behavioral experiment in which humans performed simple tasks with autonomous agents. Our results demonstrated that there were many limitations, such as a poor agent information model and trivial experimental environment. Even so, we believe our findings could lead to better research on more practical and human-friendly human-agent collaboration.

\bibliographystyle{unsrtnat}
\bibliography{reference}  






\end{document}